\documentclass[final]{cvpr}

\usepackage{graphicx}
\usepackage{amsmath}
\usepackage{amssymb}
\usepackage{booktabs}

\usepackage[pagebackref,breaklinks,colorlinks]{hyperref}

\usepackage[capitalize]{cleveref}
\crefname{section}{Sec.}{Secs.}
\Crefname{section}{Section}{Sections}
\Crefname{table}{Table}{Tables}
\crefname{table}{Tab.}{Tabs.}

\def\cvprPaperID{*****} 

\def\confYear{2023}

\begin{document}

\title{MIPI 2023 Challenge on RGB+ToF Depth Completion:\\ Methods and Results}

\author{
Qingpeng Zhu \and Wenxiu Sun \and Yuekun Dai \and Chongyi Li \and Shangchen Zhou \and Ruicheng Feng \and Qianhui Sun \and Chen Change Loy \and Jinwei Gu \and
Yi Yu\and Yangke Huang\and Kang Zhang \and
Meiya Chen\and Yu Wang\and Yongchao Li\and Hao Jiang \and
Amrit Kumar Muduli\and Vikash Kumar\and Kunal Swami\and Pankaj Kumar Bajpai \and
Yunchao Ma\and Jiajun Xiao\and Zhi Ling
}
\maketitle

\begin{abstract}
Depth completion from RGB images and sparse Time-of-Flight (ToF) measurements is an important problem in computer vision and robotics. While traditional methods for depth completion have relied on stereo vision or structured light techniques, recent advances in deep learning have enabled more accurate and efficient completion of depth maps from RGB images and sparse ToF measurements. To evaluate the performance of different depth completion methods, we organized an RGB+sparse ToF depth completion competition. The competition aimed to encourage research in this area by providing a standardized dataset and evaluation metrics to compare the accuracy of different approaches. In this report, we present the results of the competition and analyze the strengths and weaknesses of the top-performing methods. We also discuss the implications of our findings for future research in RGB+sparse ToF depth completion. We hope that this competition and report will help to advance the state-of-the-art in this important area of research. More details of this challenge and the link to the dataset can be found at \href{https://mipi-challenge.org/MIPI2023/}{https://mipi-challenge.org/MIPI2023/}.

\end{abstract}
\newcommand{\email}[1]{\href{mailto:#1}{\texttt{#1}}}
\let\thefootnote\relax\footnotetext{
\tiny Qingpeng Zhu$^{1}$(\email{zhuqingpeng@tetras.ai}), Wenxiu Sun$^{1}$(\email{sunwx@tetras.ai}), Yuekun Dai$^{2}$, Chongyi Li$^{2}$ , Shangchen Zhou$^{2}$, Ruicheng Feng$^{2}$, Qianhui Sun$^{3}$, Chen Change Loy$^{2}$, Jinwei Gu$^{1,3}$ are the MIPI 2023 challenge organizers
($^{1}$SenseTime Research and Tetras.AI, $^{2}$Nanyang Technological University, $^{3}$SenseBrain). The other authors participated in the challenge. Please refer to Appendix for details.\\
\\
MIPI 2023 challenge website: \url{https://mipi-challenge.org/MIPI2023/}
}

\section{Introduction}
\label{sec:intro}

RGB+sparse ToF depth completion is a novel approach to depth estimation in computer vision that combines RGB images with sparse depth measurements obtained from time-of-flight (ToF) sensors. Depth estimation is a fundamental problem in computer vision with numerous applications, such as robotics, autonomous driving, and augmented reality. However, estimating depth from a single modality can be challenging due to noise, occlusions, and other factors. RGB+sparse ToF depth completion has the potential to improve the accuracy and robustness of depth estimation in various real-world scenarios.

The task involves predicting a dense depth map from a single RGB image and a sparse set of depth measurements. By combining RGB images and sparse depth measurements, RGB+sparse ToF depth completion aims to leverage the complementary information provided by both modalities to produce accurate and detailed depth maps. RGB images capture color information, while the sparse depth measurements provide direct distance information about the scene. This allows for more robust and accurate depth estimation, particularly in challenging scenarios such as low-light conditions, texture-less regions, and scenes with reflective surfaces.

This challenge is a part of the Mobile Intelligent Photography and Imaging (MIPI) 2023 workshop and challenges that emphasize the integration of novel image sensors and imaging algorithms, which is held in conjunction with CVPR 2023. It consists of four competition tracks:
\begin{itemize}
  \item \textbf{Nighttime Flare Removal} is to improve nighttime image quality by removing lens flare effects.
  \item \textbf{RGB+ToF Depth Completion} uses sparse, noisy ToF depth measurements with RGB images to obtain a complete depth map.
  \item \textbf{RGBW Sensor Re-mosaic} converts RGBW RAW data into Bayer format so that it can be processed with standard ISPs.
  \item \textbf{RGBW Sensor Fusion} fuses Bayer data and monochrome channel data into Bayer format to increase SNR and spatial resolution.
\end{itemize}

\section{Challenge}
\subsection{Problem Definition}
Depth completion~\cite{lin2022dynamic,hu2022deep,hu2021penet,lee2021depth,eldesokey2020uncertainty,li2020multi,lopez2020project,cheng2020cspn++,park2020non,qu2020depth,eldesokey2019confidence,imran2019depth,cheng2019learning,chen2018estimating} aims to recover dense depth from sparse depth measurements. Earlier methods concentrate on retrieving dense depth maps only from the sparse ones. However, these approaches are limited and not able to recover depth details and semantic information without the availability of multi-modal data.
In this challenge, we focus on the RGB+ToF sensor fusion, where a pre-aligned RGB image is also available as guidance for depth completion. 
In our evaluation, the depth resolution and RGB resolution are fixed at $256 \times 192$, and the input depth map sparsity ranges from $1.0\%$ to $1.8\%$. The target of this challenge is to predict a dense depth map given the sparsity
depth map and a pre-aligned RGB image at the allowed running time constraint (please refer to Section~\ref{sec:runtime} for details).

\subsection{Dataset: TetrasRGBD \cite{sun2022mipi}}

\begin{figure*}[t]
\centering
\includegraphics[width=13cm]{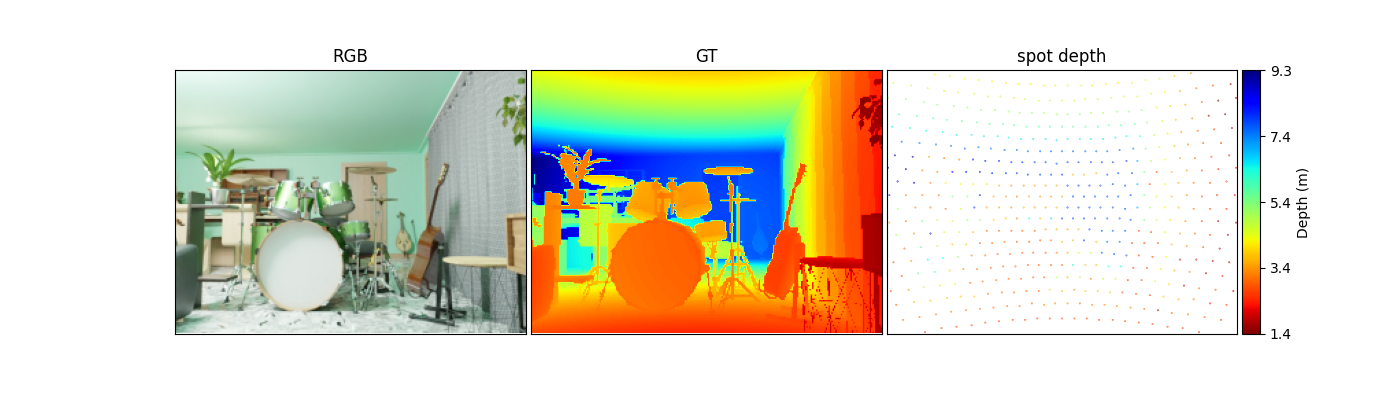}
\caption{Visualization of an example training data. Left, middle, and right images correspond to RGB, ground-truth depth, and simulated spot depth data, respectively.}
\label{fig:sample}
\setlength{\belowcaptionskip}{0pt plus 3pt minus 2pt}
\end{figure*}

The training data contains 7 image sequences of aligned RGB and ground-truth dense depth from 7 indoor scenes (20,000 pairs of RGB and depth in total). For each scene, the RGB and the ground-truth depth are rendered along a smooth trajectory in our created 3D virtual environment. RGB and dense depth images in the training set have a resolution of 640$\times$480 pixels. We also provide a function to simulate the sparse depth maps that are close to the real sensor measurements\footnote{https://github.com/zhuqingpeng/MIPI2022-RGB-ToF-depth-completion}. A visualization of an example frame of RGB, ground-truth depth, and simulated sparse depth is shown in Fig.~\ref{fig:sample}.  

The testing data contains, a) \textit{Synthetic}: a synthetic image sequence (500 pairs of RGB and depth in total) rendered from an indoor virtual environment that differs from the training data; b) \textit{iPhone dynamic}: 24 image sequences of dynamic scenes collected from an iPhone 12Pro (600 pairs of RGB and depth in total); c) \textit{iPhone static}: 24 image sequences of static scenes collected from an iPhone 12Pro (600 pairs of RGB and depth in total);  d) \textit{Modified phone static}: 24 image sequences of static scenes (600 pairs of RGB and depth in total) collected from a modified phone. Please note that depth noises, missing depth values in low reflectance regions, and mismatch of the field of views between RGB and ToF cameras could be observed from this real data. RGB and dense depth images in the entire testing set have a resolution of 256$\times$192 pixels. RGB and spot depth data from the testing set are provided and the GT depth is not available to participants. The depth data in both training and testing sets are in meters.

\subsection{Challenge Phases}
The challenge consisted of the following phases:
\begin{enumerate}
    \item Development: The registered participants get access to the data and baseline code, and are able to train the models and evaluate their running time locally.
    \item Validation: The participants can upload their models to the remote server to check the fidelity scores on the validation dataset, and to compare their results on the validation leaderboard.
    \item Testing: The participants submit their final results, code, models, and factsheets.
\end{enumerate}

\subsection{Performance Evaluation}
\subsubsection{Objective Evaluation}
We define the following metrics to evaluate the performance of depth completion algorithms.
\begin{itemize}
    \item Relative Mean Absolute Error (RMAE), which measures the relative depth error between the completed depth and the ground truth, i.e.
    \begin{equation}
    \small
    \text{RMAE} = \frac{1}{M \cdot N}\sum_{m=1}^{M}\sum_{n=1}^{N} \left\lvert\frac{\hat{D}(m,n)-D(m,n)}{D(m,n)}\right\rvert,
    \end{equation}
    where $M$ and $N$ denote the height and width of depth, respectively. $D$ and $\hat{D}$ represent the ground-truth depth and the predicted depth, respectively.

    \item Edge Weighted Mean Absolute Error (EWMAE), which is a weighted average of absolute error. Regions with larger depth discontinuity are assigned higher weights. Similar to the idea of Gradient Conduction Mean Square Error (GCMSE) \cite{lopez2017}, EWMAE applies a weighting coefficient $G(m,n)$ to the absolute error between pixel $(m,n)$ in ground-truth depth $D$ and predicted depth $\hat{D}$, i.e.
    
    \begin{equation}
    \begin{array}{l}
    \text{EWMAE} = \\
    \frac{1}{M \cdot N} \left\lvert\frac{\sum\limits_{m=1}^{M}\sum\limits_{n=1}^{N}G(m,n)\cdot[\hat{D}(m,n)-D(m,n)]}{\sum\limits_{m=1}^{M}\sum\limits_{n=1}^{N}G(m,n)}\right\rvert
    \end{array},
    \end{equation}

    where the weight coefficient $G$ is computed in the same way as in \cite{lopez2017}.
    
\end{itemize}

RMAE and EWMAE will be measured on the testing data with GT depth.
We will rank the proposed algorithms according to the score calculated by the following formula, where the coefficients are designed to balance the values of different metrics,
\begin{equation}
    \text{Objective score} = 1 - 1.8 \times \text{RMAE} - 0.6 \times \text{EWMAE}.
\end{equation}
For each dataset, we report the average results over all the processed images belonging to it.

\subsubsection{Subjective Evaluation}
For subjective evaluation, we adapt the commonly used Mean Opinion Score (MOS) with blind evaluation. The score is on a scale of 1 (bad) to 5 (excellent). We invited 13 expert observers to watch videos and give their subjective scores independently. The scores of all subjects are averaged as the final MOS.

\subsection{Running Time Evaluation}
\label{sec:runtime}
The proposed algorithms are required to be able to process the RGB and sparse depth sequence  in real time.  Participants are required to include the average run time of one pair of RGB and depth data using their algorithms and the information in the device in the submitted readme file. Due to the difference of devices for evaluation, we set different requirements of running time for different types of devices according to the AI benchmark data from the website\footnote{https://ai-benchmark.com/ranking\_deeplearning\_detailed.html}.

\section{Challenge Results}
\begin{table*}[ht]
    \centering
    \caption{MIPI 2023 RGB+ToF Depth Completion challenge results and final rankings.}
    \small
    \begin{tabular}{c|l|ll|l|l|l}
    \hline
        Rank & Team name & RMAE & EWMAE & Objective score & Subjective score & Final score  \\ \hline \hline
        1 & MGTV & 0.02164  & 0.13257  & 0.88151  & 3.60577  & 0.80133   \\ \hline
        2 & MiMcAlgo & 0.01921  & 0.13462  & 0.88465  & 3.38846  & 0.78117   \\ \hline
        3 & Dintel & 0.02702  & 0.13569  & 0.86995  & 3.06731  & 0.74171   \\ \hline
        4 & Chameleon & 0.02232  & 0.13305  & 0.87999  & 2.90385  & 0.73038   \\ \hline
    \end{tabular}
    \label{tab:results}
\end{table*}

Among $77$ registered participants, $4$ teams successfully submitted their results, code, and factsheets in the final test phase. Table~\ref{tab:results} summarizes the final test results and rankings of the teams. Team MGTV shows the best overall performance, followed by Team MiMcAlgo and Team DIntel. The proposed solutions are described in Section \ref{sec:methods} and the team members and affiliations are listed in Appendix \ref{append:teams}.

\section{Challenge Methods}
\label{sec:methods}
\subsection{MGTV}

\begin{figure}[ht]
    \centering
    \includegraphics[width=8cm]{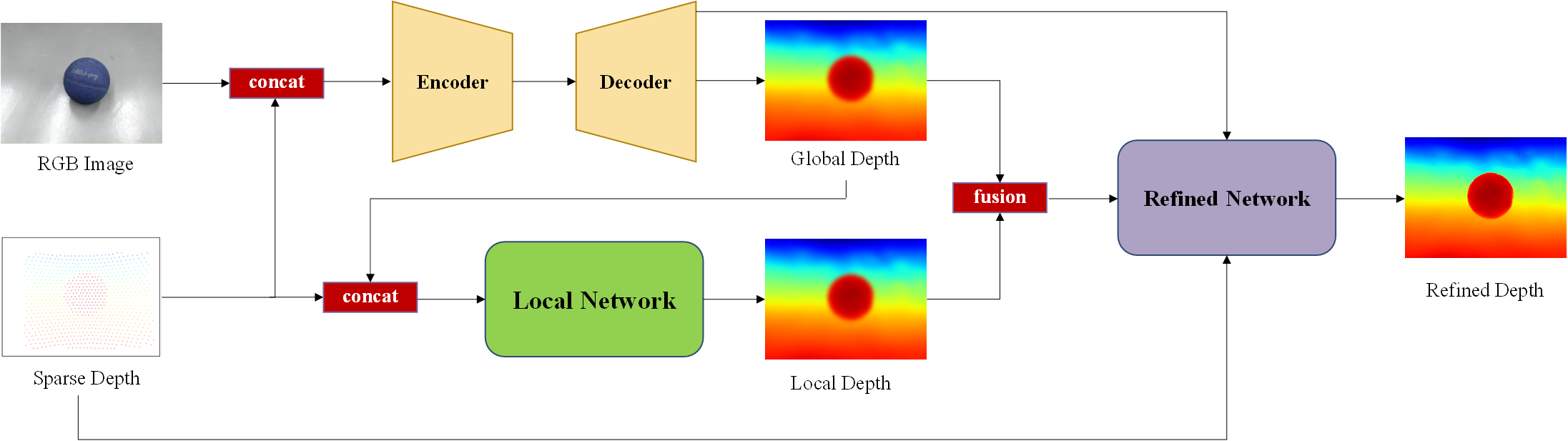}
    \caption{Framework of the MGTV team}
    \label{fig:mgtv}
\end{figure}

\noindent \textbf{Model.}This team has proposed a model architecture, as illustrated in Figure \ref{fig:mgtv}, that incorporates ResNet34~\cite{he2016deep} for the Encoder module, while the Decoder module is composed of a sequence of 5 upsample blocks, each of which integrates a convolutional layer, a pixel shuffle layer, and a channel attention branch~\cite{hu2018squeeze}. The Local Network is a concatenation of three consecutive convolutional layers, while the Refined Network capitalizes on the FCSPNet~\cite{hou2023learning} for optimal performance.

\noindent \textbf{Loss.}This team employed three loss functions jointly for training, namely L1 loss, RMAE loss, and Gradient loss proposed by ~\cite{hou2023learning}. In the early stage, in order to ensure the stability of model training, the L1 losses were calculated for global depth, local depth, and refined depth simultaneously. After the model training became stable, only the L1 loss for refined depth was calculated to obtain optimal results.

\noindent \textbf{Data augmentation.}This team observed that the sparse depth inputs in the testing data were not all sampled on a regular grid. During training, This team randomly applied local and circular removal operations to the sparse depth inputs with a certain probability, to simulate the input distribution of sparse depth data in the testing phase. This approach not only enhanced the generalization capability of the model but also improved its performance on such testing data.

\noindent \textbf{Outlier handling.} 
This team observed that the sparse depth inputs in the testing data have some outliers which will introduce bias in the predictions around them. So this team removed the outliers beyond 3 standard deviations from the mean during the inference phase.

\subsection{MiMcAlgo}
\begin{figure}[ht]
    \centering
    \includegraphics[scale=0.4]{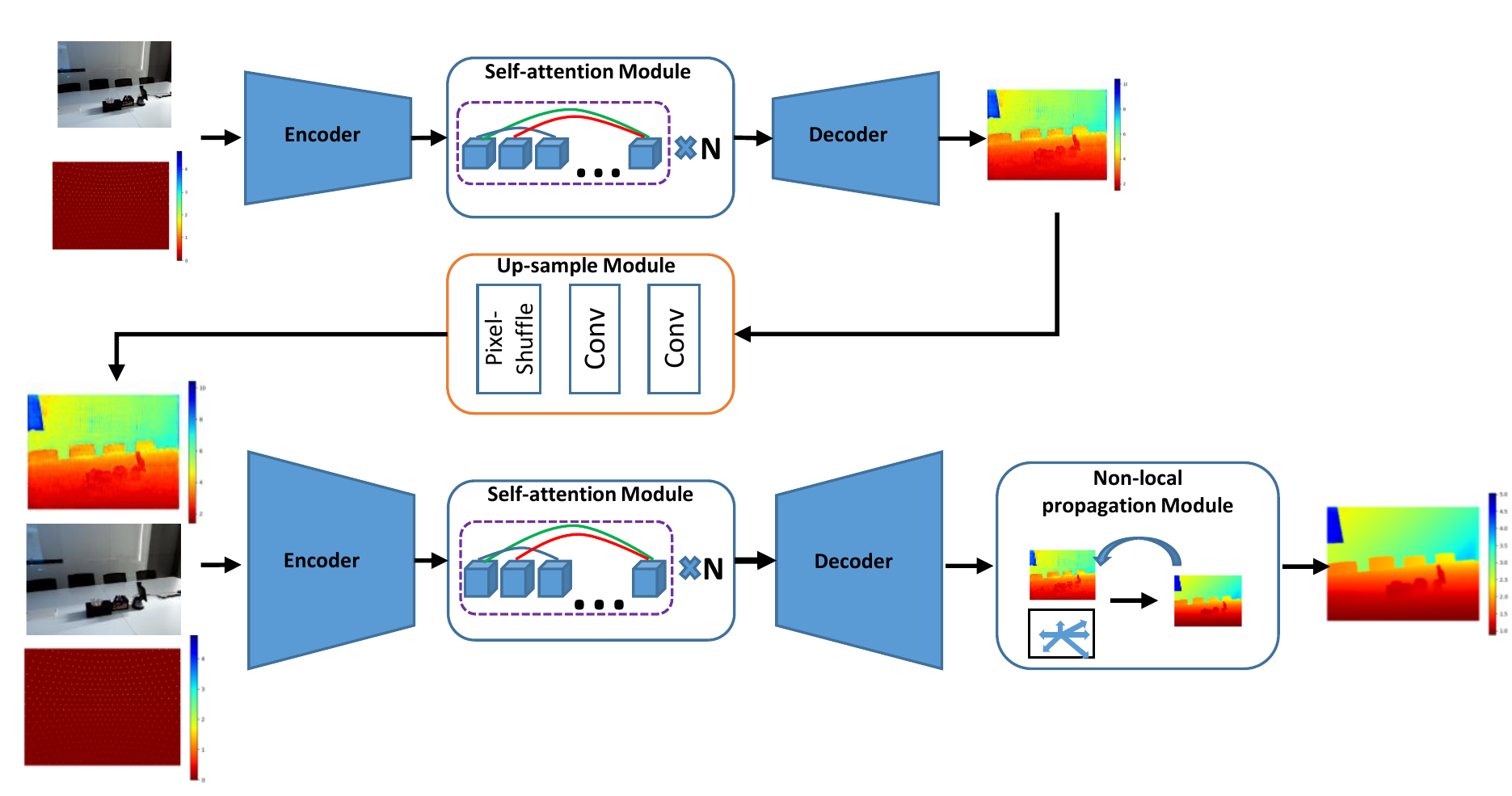}
    \caption{Overview of EMNLP-TN.}
    \label{fig_EMNLP-TN}
\end{figure}

\noindent \textbf{Method.} The team has developed an effective and efficient method for depth completion called Enhancing Multi-Scale Non-Local Propagation with Transformer Network (EMNLP-TN), illustrated in Figure \ref{fig_EMNLP-TN}.
\newline {Coarse-to-fine prediction:} The proposed method, EMNLP-TN, employs a two-stage approach for depth completion, which shares similarities with the method outlined in penet \cite{hu2021penet}. However, EMNLP-TN utilizes a coarse-to-fine framework that addresses the complexity of depth completion in two sub-problems: structure prediction and detail refinement. In both stages, the same encoder and decoder network are utilized. Specifically, the second stage takes the coarse depth prediction from the first stage as input and applies several non-local spatial propagation networks \cite{park2020non} to enhance the prediction.
\newline {Artifacts-free up-sample:} From the perspective of super-resolution, completion can also be seen as the process of interpolating missing depth information using RGB image guidance. This makes depth completion an equivalent process to depth super-resolution. To achieve this, the team used the pixel-shuffle\cite{aitken2017checkerboard} operation from super-resolution to bridge the gap between downscaled and full-sized depth data. 
\begin{figure}[ht]
	\centerline{\includegraphics[scale=0.3]{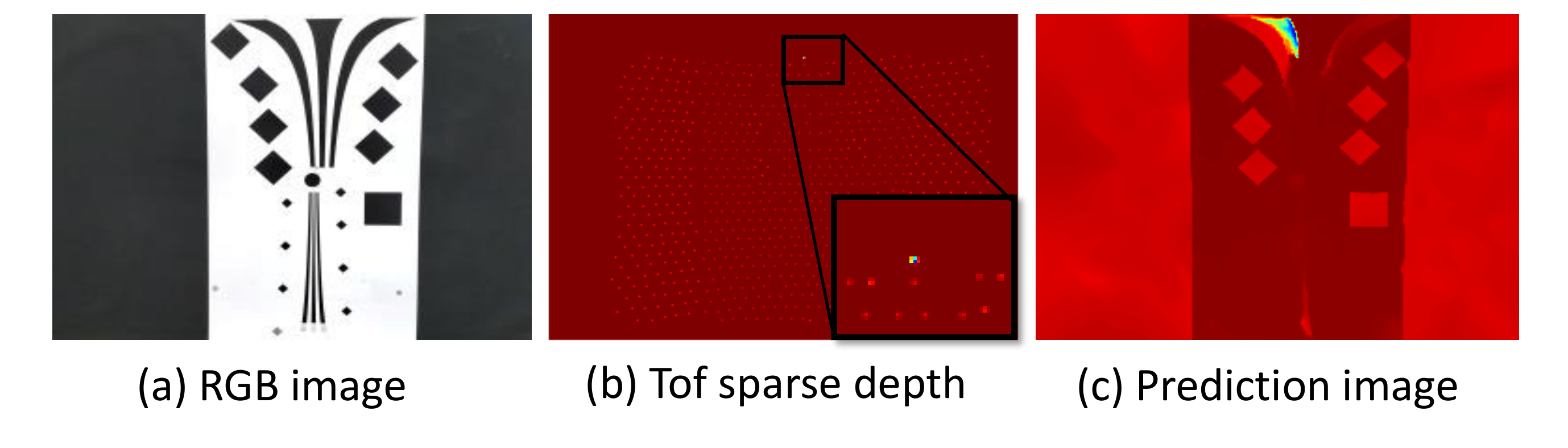}}
	\caption{Bad point result.}
	\label{bad_point}
\end{figure}
\newline {Self-attention:} Depth completion is distinct from stereo depth because it does not rely on depth cues from the input. Instead, it leverages sparse depth information to label segments or pixels and learns semantic and segmentation information from RGB images. As demonstrated by the example illustrated in Figure \ref{bad_point}, the flat input image can yield widely varying depths in different segments due to shortcomings in the sparse depth map. The results confirm that the network learns more segmentation information rather than depth. This indicates that the network is learning more about segmentation than depth. To address this issue, the team employed a transformer\cite{vaswani2017attention} to extract global features for better segmentation. The self-attention module of light LoFTR\cite{sun2021loftr} is introduced to aggregate more sparse depth points. 

\noindent \textbf{Training loss.} Each branch is trained with
\begin{equation}
    Loss = \alpha * RMAE+\beta * EWMAE.
    \label{eqn:loss-MiMcAlgo1}
\end{equation}
Specifically, the loss shares the same formulation as RMAE and EWMAE metrics.
\newline The total loss of EMNLP-TN can be described as follow
\begin{equation}
Loss_{total} = Loss_{corase}+Loss_{fine}.
\end{equation}

\subsection{DIntel}
\begin{figure}[ht]
    \centering
    \includegraphics[scale=0.23]{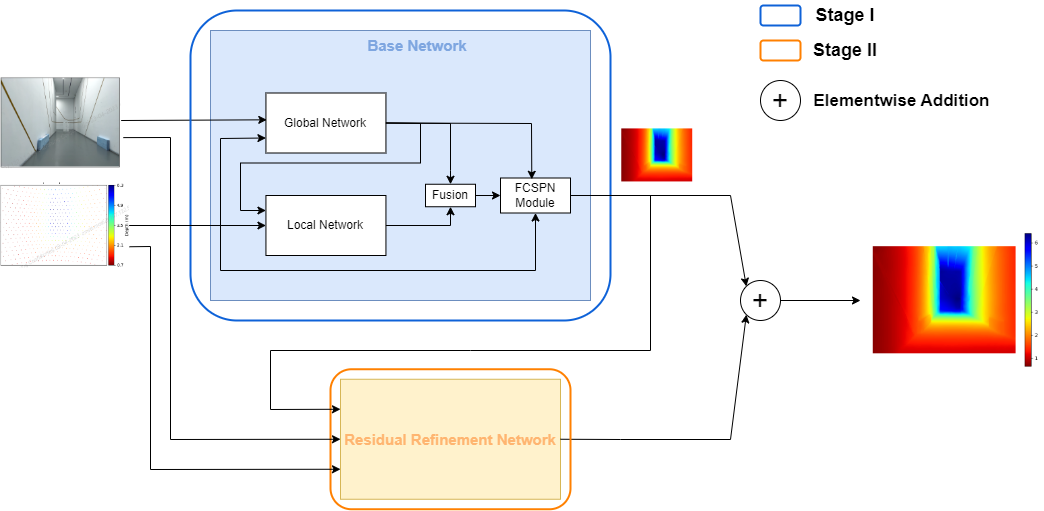}
    \caption{Overview of DIntel architecture}
    \label{fig_DIntel}
\end{figure}

\noindent \textbf{Model.} This team uses a cascaded residual refinement approach for the depth completion task. In this cascaded style network architecture, the authors have used two network modules: Base Network and Residual Refinement Network (RRN), as shown in Figure \ref{fig_DIntel}. Inspired from \cite{hou2023learning}, the base network's architecture gives local and global depth predictions, which are then fused and passed through a few more layers to get the affinity matrix for funnel convolutional spatial propagation network (FCSPN) module, which in turn gives the initial refined depth map prediction. The output of the base network, along with RGB image and spot depth, is then fed as input to the RRN (a ResNet-34 \cite{he2016deep} based architecture), whose objective is to rectify the incorrect depth predictions from the base network. 

\noindent \textbf{Training.} The base network was trained similarly to \cite{hou2023learning}. After training the base network, the weights of the base network were frozen, and only the RRN was trained. RRN was trained for 42 epochs with a batch size of 16. The initial learning rate was 0.001 with a decay factor of \{1.0, 0.2, 0.04\} at epochs \{10, 15, 20\}. To tackle the missing sparse depth challenge in the test dataset, the authors trained the model by removing the spot depth from random areas of the image.

\noindent \textbf{Loss.} In the base network and RRN, the authors have used a combination of $L_{1}$, $L_{2}$ and Corrected Gradient Loss (CGDL) \cite{hou2023learning} with adaptive weights of 1, 1 and 0.7, respectively. While $L_{1}$ and $L_{2}$ losses helped get accurate depths, the main objective of CGDL loss was to keep the depth prediction output smooth and edges sharp.

\subsection{Chameleon}
\begin{figure}
    \centering
    \includegraphics[width=8cm]{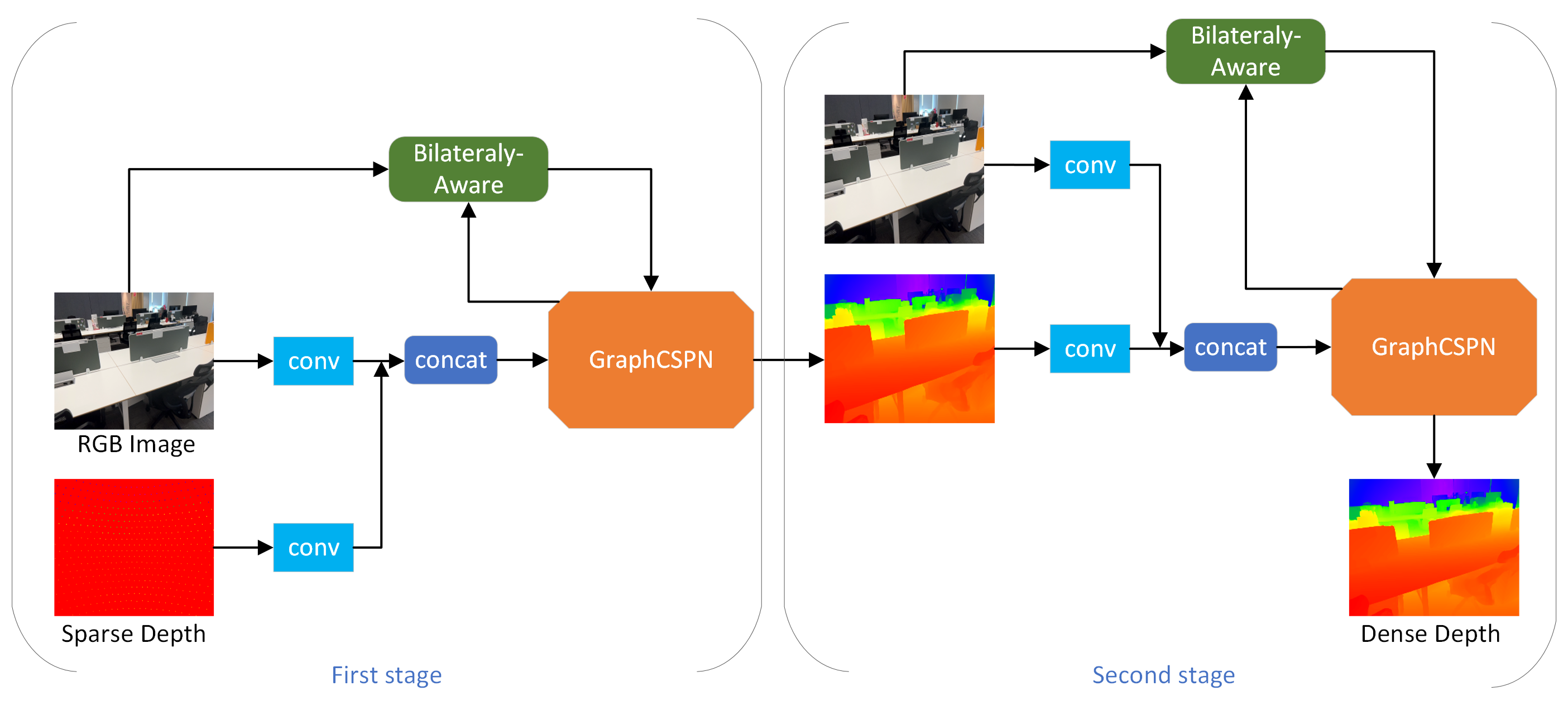}
    \caption{Pipeline of CDGNN-BA.}
    \label{fig_CDGNN-BA}
\end{figure}

\noindent \textbf{Model.} This team propose an effective and efficient depth completion method for depth completion called Bilateraly-Aware Depth Completion via Cascaded Dynamic Grapth Convolutional Network (CDGNN-BA), illustrated in Figure \ref{fig_CDGNN-BA}.
The proposed method, CDGNN-BA, outputs dense depth and follows the Convolutional Spatial Propagation Network, originally introduced in \cite {cheng2019learning}. In particular, this team start out with Dynamic Grapth Convolutional Network design as in GraphCSPN \cite{liu2022graphcspn} and propose modifications for further improvement as below.
\newline {Bilateraly-Aware:} GraphCSPN utilizes an encoder-decoder to jointly learn the initial depth map and affinity matrix, which is sampled and reshaped into sequences of patches and concatenated with 3D position embeddings. Then the model estimates the neighbors of different patches on the basis of geometric constraints, and performs spatial propagation leveraging dynamic graph convolution networks with self-attention mechanism. After the graph is generated, the Top k patches are judged, and the weighted sum is obtained to predict the final depth. However, the way of calculating the similarity only considers the physical distance between patches while the heterochromia from various objects is ignored. In our method, the RGB information is considered based on the fact that patches from the same object have a higher correlation; Additionally, the naive 3D reconstruction requires camera internal parameters which are hard to obtain in several scenarios. Instead of using these parameters, the authors utilized the distances between pixels and depth distance from the depth map. In particular, the above distance weights are calculated using bilateral weight:
\begin{equation}
d = e^{(-\frac{(d_{rgb} - u_{rgb})^2}{2\sigma_{rgb}^2} - \frac{(d_{pixel} - u_{pixel})^2}{2\sigma_{pixel}^2} - \frac{(d_{depth} - u_{depth})^2}{2\sigma_{depth}^2})}.
\end{equation}
\newline {Cascaded refinement:} GraphCSPN re-samples the affinity matrix by 3-time losses the information of original features. For compensating this deficiency, the cascade structure is proposed. First, 4-time down-sampling is performed. After graph propagation, coarse depth is obtained, and then 2-time down-sampling is applied to obtain refined depth.

\noindent \textbf{Loss.} This team uses the same loss function as Equation \ref{eqn:loss-MiMcAlgo1}

\noindent \textbf{Training Setup.} For training, this team randomly cropped and resized 192 × 256 patches from the training images as inputs. The mini-batch size is set to 4 and the whole network is trained for 200 epochs. The learning rate is initialized as $1\times10^{-3}$, decayed with a LinearLR schedule.

\section{Conclusions}
In this report, we review and summarize the methods and results of MIPI 2023
challenge on RGB+ToF Depth Completion. The participants were provided with a high-quality training/testing dataset, TetrasRGBD, which is now available for researchers to download for future research. We are excited to see the new progress contributed by the submitted solutions in such a short time, which are all described in this paper. The challenge results are reported and analyzed. Detailed descriptions of the submitted solutions are also provided in this report. For future works, there is still plenty of room for improvements including dealing with depth outliers/noises, precise depth boundaries, high depth resolution, and high temporal stability, etc.

\section{Acknowledgements}
We thank Shanghai Artificial Intelligence Laboratory, Sony, Nanyang Technological University and The Chinese University of Hong Kong to sponsor this MIPI 2023 challenge. We thank all the organizers for their contributions to this workshop and all the participants for their great work.

{\small
\bibliographystyle{ieee_fullname}
\bibliography{mybib}
}

\appendix

\section{Teams and Affiliations}
\label{append:teams}

\subsection*{\bf MGTV team}
\noindent
{\bf Title:} MangoTV\\
{\bf Members:}\\
YiYu$^1$ (\href{yuyi@mgtv.com}{yuyi@mgtv.com})\\
YangkeHuang$^1$(\href{yangke5@mgtv.com}{yangke5@mgtv.com})\quad KangZhang$^1$(\href{zhangkang@mgtv.com}{zhangkang@mgtv.com})\\
{\bf Affiliations:}\\
$^1$ MGTV (\href{https://w.mgtv.com/}{https://w.mgtv.com/})

\subsection*{\bf MiMcAlgo team}
\noindent
{\bf Title:} Enhancing Multi-Scale Non-Local Propagation with Transformer Network.

\noindent
{\bf Members:} \\
Meiya Chen (\href{chenmeiya@xiaomi.com}{chenmeiya@xiaomi.com}) \\
Yu Wang (\href{wangyu50@xiaomi.com}{wangyu50@xiaomi.com}) \\
Yongchao Li (\href{liyongchao1@xiaomi.com}{liyongchao1@xiaomi.com}) \\
Hao Jiang (\href{jianghao11@xiaomi.com}{jianghao11@xiaomi.com})

\noindent
{\bf Affiliations:} \\
Xiaomi Inc.

\subsection*{\bf DIntel team}
\noindent
{\bf Title:} Efficient and Refined Multi-modal Depth Completion\\
\noindent
{\bf Members:} \\
Amrit Kumar Muduli (\href{amrit.muduli@samsung.com}{amrit.muduli@samsung.com})\\
Vikash Kumar (\href{vikash.k7@samsung.com}{vikash.k7@samsung.com})\\
Kunal Swami (\href{kunal.swami@samsung.com}{kunal.swami@samsung.com})\\
Pankaj Kumar Bajpai (\href{pankaj.b@samsung.com}{pankaj.b@samsung.com})\\
\noindent
{\bf Affiliation:} \\
Samsung R \& D Institute India - Bangalore

\subsection*{\bf Chameleon team}
\noindent
{\bf Title:} Bilateraly-Aware Depth Completion via Cascaded Dynamic Grapth Convolutional Network

\noindent
{\bf Members:} \\
Yunchao Ma (\href{mayunchao@megvii.com}{mayunchao@megvii.com})\\
Jiaojun Xiao (\href{xiaojiajun@megvii.com}{xiaojiajun@megvii.com})\\
Zhi Ling (\href{lingzhi@megvii.com}{lingzhi@megvii.com})\\

\end{document}